\pdfoutput=1

\documentclass[11pt]{article}

\usepackage[final]{acl}

\usepackage{times}
\usepackage{enumitem}
\setlist{topsep=0pt, leftmargin=*}
\usepackage{latexsym}
\usepackage{amsmath}
\usepackage{amssymb}
\usepackage{enumerate}
\usepackage{tabularray}
\usepackage{algorithm}
\usepackage{algorithmic}
\usepackage[symbol]{footmisc}
\usepackage{xcolor}
\usepackage{hyperref}
\hypersetup{colorlinks=true,linkcolor=blue,urlcolor=blue}

\usepackage[T1]{fontenc}

\usepackage[utf8]{inputenc}

\usepackage{microtype}

\usepackage{inconsolata}

\usepackage{graphicx}

\definecolor{bluegray}{rgb}{0.4, 0.6, 0.8}

\title{DocTalk: Scalable Graph-based Dialogue Synthesis \\for Enhancing LLM Conversational Capabilities}

\author{Jing Yang Lee\textsuperscript{1}\footnotemark, Hamed Bonab\textsuperscript{2}, Nasser Zalmout\textsuperscript{2}, Ming Zeng\textsuperscript{2}, Sanket Lokegaonkar\textsuperscript{2}\\ \bf Colin Lockard\textsuperscript{2},  Binxuan Huang\textsuperscript{2}, Ritesh Sarkhel\textsuperscript{2}, Haodong Wang\textsuperscript{2} \\
  Nanyang Technological University\textsuperscript{1},
  Amazon\textsuperscript{2}\\
  jingyang001@e.ntu.edu.sg,  \{hamedrab, nzalmout, minzen, sllokega\}@amazon.com\\\{clockard, binxuan, rssarkhe, wanghaod\}@amazon.com}

\begin{document}
\maketitle
\footnotetext[1]{Work done during internship at Amazon.}
\begin{abstract}
Large Language Models (LLMs) are increasingly employed in multi-turn conversational tasks, yet their pre-training data predominantly consists of continuous prose, creating a potential mismatch between required capabilities and training paradigms. We introduce a novel approach to address this discrepancy by synthesizing conversational data from existing text corpora. We present a pipeline that transforms a cluster of multiple related documents into an extended multi-turn, multi-topic information-seeking dialogue. Applying our pipeline to Wikipedia articles, we curate \emph{DocTalk}, a multi-turn pre-training dialogue corpus consisting of over $730$k long conversations. We hypothesize that exposure to such synthesized conversational structures during pre-training can enhance the fundamental multi-turn capabilities of LLMs, such as context memory and understanding. Empirically, we show that incorporating \emph{DocTalk} during pre-training results in up to $40$\% gain in context memory and understanding, without compromising base performance. \emph{DocTalk} is available at \url{https://huggingface.co/datasets/AmazonScience/DocTalk}.

\end{abstract}

\section{Introduction}
Users expect conversational AI assistants to deliver dynamic and natural dialogue, driving a significant increase in Large Language Model (LLM) deployment. These multi-turn, multi-topic conversations involve users asking questions and LLMs providing relevant answers while maintaining context and adapting to topic shifts. However, LLMs are primarily pre-trained on large datasets of prosaic web-sourced text, including news, books, and papers with only a small portion comprising conversational data like chat logs or forums \cite{longpre2024pretrainer,zhenglmsys}. The limited amount of conversational data creates a gap between pre-training and typical LLM usage. Existing dialogue datasets largely focus on single-topic conversations, unlike real-world interactions, which span 1 to 10 topics, averaging 2 per session \cite{spink}. 

By bridging this gap during pre-training, we aim to enhance the LLM's fundamental multi-turn capabilities, particularly in context memory and understanding, crucial for maintaining coherence and relevance in extended dialogues \cite{Bai_2024}. Context memory refers to the LLM's ability to accurately retrieve and utilize prior dialogue information to ensure continuity and relevance. Context understanding, on the other hand, involves resolving anaphora (e.g., identifying pronoun referents) and linking instructions to subsequent inputs across multiple dialogue turns for effective task comprehension. Emphasizing these capabilities during pre-training lays a robust foundation for advanced multi-turn reasoning in fine-tuning.


We propose a conversational data synthesis pipeline that converts web-derived prose into multi-turn, multi-topic, information-seeking dialogues featuring multiple topic switches. Unlike prior approaches that rely heavily on LLMs to generate a large portion of the data \cite{xu2024magpiealignmentdatasynthesis,nayak2024learninggenerateinstructiontuning, ge2024scalingsyntheticdatacreation, long-etal-2024-llms}, our pipeline minimizes the use of LLM-generated text, reducing the risk of hallucination in addition to improving the quality and reliability of synthesized dialogues \cite{liu2024bestpracticeslessonslearned}. Our approach also provides substantial cost benefits, achieving a 70\% reduction in generation costs ( Appendix \ref{sec:cost}), significantly enhancing scalability. In contrast to methods like Dialogue Inpainting \cite{dai2022dialoginpaintingturningdocuments}, which produces short, single-topic conversations, our pipeline models the dynamic topical shifts observed in human-LLM interactions, resulting in long, multi-topic conversations.



Applying this pipeline to Wikipedia, we create \textit{DocTalk}, the largest multi-turn conversational dataset by token count to date, to our knowledge. We continue pre-training Mistral-7B-v0.3 on \textit{DocTalk} and evaluate context memory and understanding using the Conversational Question Answering (CoQA) corpus and a novel LLM-as-a-judge framework. These metrics assess turn-level performance and the ability to incorporate relevant context in extended dialogues. We also employ guardrail metrics to investigate any potential degradation in existing LLM capabilities. Our experiments show that continued pre-training on data mixtures consisting of multi-turn, multi-topic information-seeking conversational data significantly enhances context memory and understanding, without compromising existing capabilities. Our key contributions are as follows:
\begin{itemize}[noitemsep,topsep=-8pt]
\item We introduce a scalable pipeline for synthesizing conversational data that transforms multiple documents into coherent multi-turn, multi-topic information-seeking dialogues.
\item We create \textit{DocTalk}, the largest conversational dataset by token count, by applying our pipeline to Wikipedia documents, demonstrating its effectiveness at scale.
\item We provide empirical evidence that our approach enhances LLM context memory and understanding while preserving other model capabilities.
\end{itemize}

\section{Methodology}
Our conversational data synthesis pipeline consists of three stages. In \textit{Stage 1}, we construct a document graph connecting multiple related documents and sample a set of documents from it. Leveraging multiple documents, rather than a single source, enables the creation of conversations that span multiple topics. In \textit{Stage 2}, we divide the content of the sampled documents into segments and build a dialogue graph modeling probabilities of switching from one segment to another in natural conversations. We then sample these segments in sequence according to the dialogue graph and consider them as ``assistant'' utterances. Finally in \textit{Stage 3}, we generate the corresponding user utterances by prompting an off-the-shelf LLM model to provide questions eliciting the assistant's responses at each conversational turn. We posit that pre-training on such structured data improves dialogue capabilities, particularly in aspects of contextual memory and comprehension.

\subsection{Stage 1: Document Graph (GDoc)}
We aim to systematically obtain and compile a set of $n$ related documents to synthesize structured multi-turn dialogues. This process requires a pool of interconnected documents, where interconnections can be established through various meaningful dimensions of relatedness. Here, we primarily focus on explicit connections through direct references between documents \cite{frej-etal-2020-wikir}. Alternative relatedness, such as sequential or hierarchical relationships and semantic topical similarity \cite{shi2024incontextpretraininglanguagemodeling}, remain as potential directions for future exploration.

\begin{figure}[]
    \centering
    \scalebox{0.55}{
    \includegraphics[]{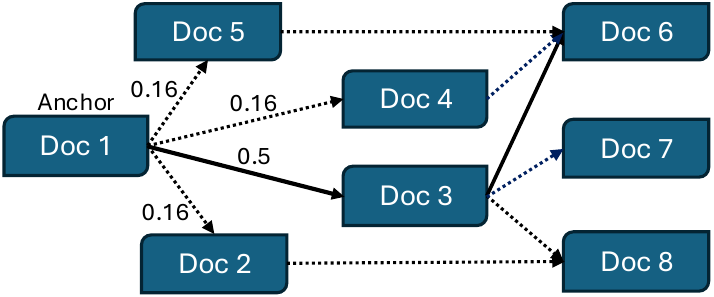}}
    \caption{An illustrative example of GDoc constructed in \emph{Stage 1}. The edges (dotted arrows) originating from Doc 1 (the ``anchor'') are labeled with their respective sampling probabilities--based on the out-degree centrality. The bold arrows indicate the sampled edges.}
    \label{fig:doc_graph}
\end{figure}

\noindent\textbf{GDoc Construction.} We construct a weighted directional acyclic graph, $G_{doc}=(V_{doc},E_{doc})$, where $V_{doc}$ refers to a set of documents as vertices and $E_{doc}$ refers to their connecting edges, i.e., $E_{doc}  \subseteq \{(v_{doc}^{i},v_{doc}^{j}) \vert v_{doc}^{i},v_{doc}^{j} \in V_{doc},i \neq j\}$. Our document graph construction starts with a random ``anchor'' document. Then, all related documents are collated to form the first level of connected vertices. Subsequently, from each of the newly formed vertices, related documents will be collated to form the next level of the graph--we stop at depth=3. For assigning edge weights, we adopt a topological approach based on vertex centrality measurements. For edge $e_{i,j}$ connecting $v_{i}$ to $v_{j}$, the edge weight $w_{{i,j}} = deg^{\text{-}}(v_{j})$ where $deg^{\text{-}}$ refers to the out-degree centrality \cite{xamena2017structural}. Out-degree centrality measures a document's direct references to others, indicating its prominence within the graph \cite{Yang_2014}. Also, high quality and information rich documents typically have broad coverage, referencing numerous subtopics or related articles \cite{10.1145/3442442.3452345}. This would encourage the synthesis of longer and more informative conversations. Constructing GDoc in this manner enables us to model various potential topical flows from the "anchor" document, representing possible dialogue trajectories in a conversational context. See Figure \ref{fig:doc_graph} for an illustration.

\noindent\textbf{GDoc Traversal.} We aim to sample a set of $n$ documents--we use $n=3$. Traversal begins at the anchor document, following a probabilistic random walk, where the next vertex is selected based on the associated edge weights. The traversal stops either when $n$ documents are collected or when a vertex with no outgoing edges is reached, yielding a cohesive and thematically related set of documents--see Appendix \ref{sec:algos}.

\subsection{Stage 2: Dialogue Graph (GDial)}
We construct $m$ assistant utterances from the $n$ related documents obtained in Stage 1. We first segment each document and then systematically reorder and interleave these segments to create a natural topic progression. $m$ corresponds to the total number of segments derived from $n$ documents. This interleaving of segments from different but related documents enables the creation of conversations that are both multi-topic and multi-turn. Each utterance contributes novel information while maintaining contextual references to previous utterances derived from the same source document, thereby establishing cross-turn coreference.

\begin{figure}[]
    \centering
    \scalebox{0.5}{
    \includegraphics[]{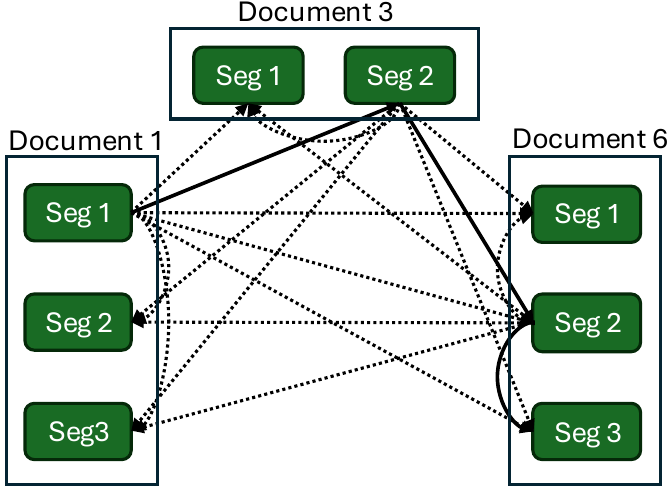}}
    \caption{An illustrative example of GDial constructed in \emph{Stage 2} assuming Doc 1, 3 and 6 are sampled during \emph{Stage 1}. Arrows represent edges and bold arrows indicate the order of 4 segments sampled for assistant utterances--i.e., [Doc1, Seg1]$\rightarrow$[Doc3, Seg2]$\rightarrow$[Doc6, Seg2 ]$\rightarrow$[Doc6, Seg3]. }
    \label{fig:dial_graph}
\end{figure}

\noindent\textbf{GDial Construction.} Using $m$ text segments extracted from documents sampled in Stage 1, we construct a separate directed graph $G_{dial}=(V_{dial},E_{dial})$. We begin by splitting each document into distinct segments, serving as vertices  $V_{dial}$, representing a potential assistant utterance. The graph is fully connected, with edges linking each vertex to every other vertex, regardless of their source documents and with no self-loops, such that $E_{dial}  = \{(v_{dial}^{i},v_{dial}^{j}) \vert v_{dial}^{i},v_{dial}^{j} \in V_{dial},i \neq j\}$. Each segment is constrained to single use, ensuring that the resulting conversations contain no duplicate assistant utterances. See Figure \ref{fig:dial_graph} for an illustration. 
A critical component in the construction of GDial is the assignment of edge weights that model natural conversational flow. We introduce a novel Conversational Reward (CR) model fine-tuned to assign scores based on the coherence between the assistant's previous utterance(s) and potential subsequent utterances represented by connected vertices. These edges are then sampled based on probabilities proportional to their assigned weights. A higher CR model score indicates that the utterance tied to the connected vertex is better suited as the assistant’s next response in a multi-turn, information-seeking conversation. This approach enables us to construct a complete graph, with the CR model guiding its traversal.

\noindent\textbf{Conversational Reward (CR) Model.} We conceptualize the CR model as a learned scoring mechanism that quantifies the appropriateness of candidate utterances in the context of preceding assistant turns within multi-turn dialogues. Drawing from established semantic matching objectives, the model is trained to evaluate local coherence and topical progression across turns, thereby assigning scalar values that reflect the conversational flow in information-seeking interactions. We formulate the objective of the CR model as estimating the conditional probability $P(A_{t+1}^{c} | A_{t})$, where $A_{t+1}^{c}$ is the candidate utterance in turn $t+1$ and $A_{t}$ is the preceding assistant utterance at turn $t$. For tractability, we restrict contextual input to one preceding assistant utterance. While this design choice facilitates computational efficiency and interpretability, extending the model to condition on broader dialogue history remains a promising direction for future research.

Our CR model builds upon a pre-trained ranking model, acknowledging relevance as a key determinant of conversational flow. We enhance this foundation through additional training on high-quality conversational data to capture dialogue-specific characteristics. We collect our training data using $4,240$ conversations aggregated from six corpora--for information seeking: HybriDialogue \cite{nakamura-etal-2022-hybridialogue}, InScitic \cite{wu-etal-2023-inscit}, TopiOCQA \cite{adlakha2022topiocqaopendomainconversationalquestion}, and Wizard of Wikipedia \cite{dinan2019wizardwikipediaknowledgepoweredconversational}; for LLM-based: ShareGPT\footnote{The ShareGPT dataset: https://sharegpt.com/}, and UltraChat \cite{ding2023enhancingchatlanguagemodels}. More details are provided in Appendix \ref{sec:cr_model}.

We use \textit{bge-reranker-base} \cite{bge_embedding} as the starting base model. To prepare training samples, we randomly select a base assistant utterance (excluding the last utterance) from each conversation in the training set, and the next assistant utterance as the ``positive'' sample $y_{p}$. For ``negative'' sample $y_{n}$, we randomly select from a $3*k$ set collected by: \textit{(a)} using the base model, get top-$k$ assistant utterances within the same conversation--as a form of hard negative mining \cite{robinson2021contrastive}; \textit{(b)} using the base model, get top-$k$ utterances from other dialogues; and \textit{(c)} Another $k$ random utterances from other dialogues. We fine-tune the base model iteratively via a pairwise log probability contrastive loss, $-log(R(y_{p})-R(y_{n}))$, where $R(\cdot)$ refers to the CR model output. We update mined hard-negatives iteratively with three iterations of the base CR model, and train each iteration for two epochs--with LR$=3e^{-5}$. 

To evaluate the CR model performance, we select $100$ conversations from each dataset's test set, with performance measured by Mean Reciprocal Rank (MRR). We train with three iterations, each with a distinct dataset corresponding to $k$ = 1, 2 and 3, respectively. Results are reported in Table \ref{tab:crmodel_performance}. We observe that finetuning with $k$ = 2 yielded the highest MRR scores for LLM-based dialogue and performed comparably to the model fine-tuned with $k$ = 3 for information-seeking dialogue. Notably, the fine-tuned models showed significant improvement in the next utterance prediction task compared to the base model. The complexity of this task presents major opportunities, suggesting multiple avenues for future investigation and enhancement.


\begin{table}
\centering
\caption{CR model performance (Averaged MRR) }
\label{tab:crmodel_performance}
\scalebox{0.8}{
\begin{tblr}{
  column{2} = {c},
  hline{1-2,6} = {-}{},
  stretch=0, colsep  = 3.0pt, rowsep=2pt
}
\textbf{Model}    & \textbf{LLM-based} & \textbf{Info Seeking} \\
bge-reranker-base & 0.199              & 0.121                 \\
Iteration 1 (w. $k=1$)           & 0.381              & 0.182                 \\
Iteration 2 (w. $k=2$)          & \textbf{0.413}     & 0.168                 \\
Iteration 3 (w. $k=3$)         & 0.371              & \textbf{0.17}         
\end{tblr}}
\end{table}

\noindent\textbf{GDial Traversal} After constructing $G_{dial}$, we sample a set of $m$ assistant utterances by performing a probabilistic random walk. At each node, the next node is sampled with a probability proportional to the edge weights assigned by the CR model. To avoid repetition, each visited vertex is excluded from future selections during the traversal. The traversal ends when every node in GDial has been visited. It should be highlighted that different paths through GDoc and GDial would yield distinct topical trajectories and unique dialogues, effectively capturing the one-to-many nature of conversational dynamics. An algorithm is provided in Appendix \ref{sec:algos}.

\subsection{Stage 3: User Utterance Generation}
We generate user utterances that precede each of the $m$ assistant utterances by prompting an LLM. Essentially, we prompt the Mistral-2-7b-Instruct to generate a question which elicits each assistant utterance in the conversation. Only the user utterances are generated in this process, and no LLM-generated content is included in the assistant utterances. While this approach may slightly impact the naturalness, it significantly minimizes hallucinations in the resulting conversations, which is critical for effective pre-training \cite{liu2024bestpracticeslessonslearned}. Our simplified approach achieves at least 70\% reduction in generation costs, significantly enhancing its potential for large-scale deployment.
Further details are provided in Appendix \ref{sec:cost} and \ref{sec:user-utt-prompt}.




\section{DocTalk}
\label{sec:doctalk}

\begin{figure}[]
    \centering
    \scalebox{0.425}{
    \includegraphics[]{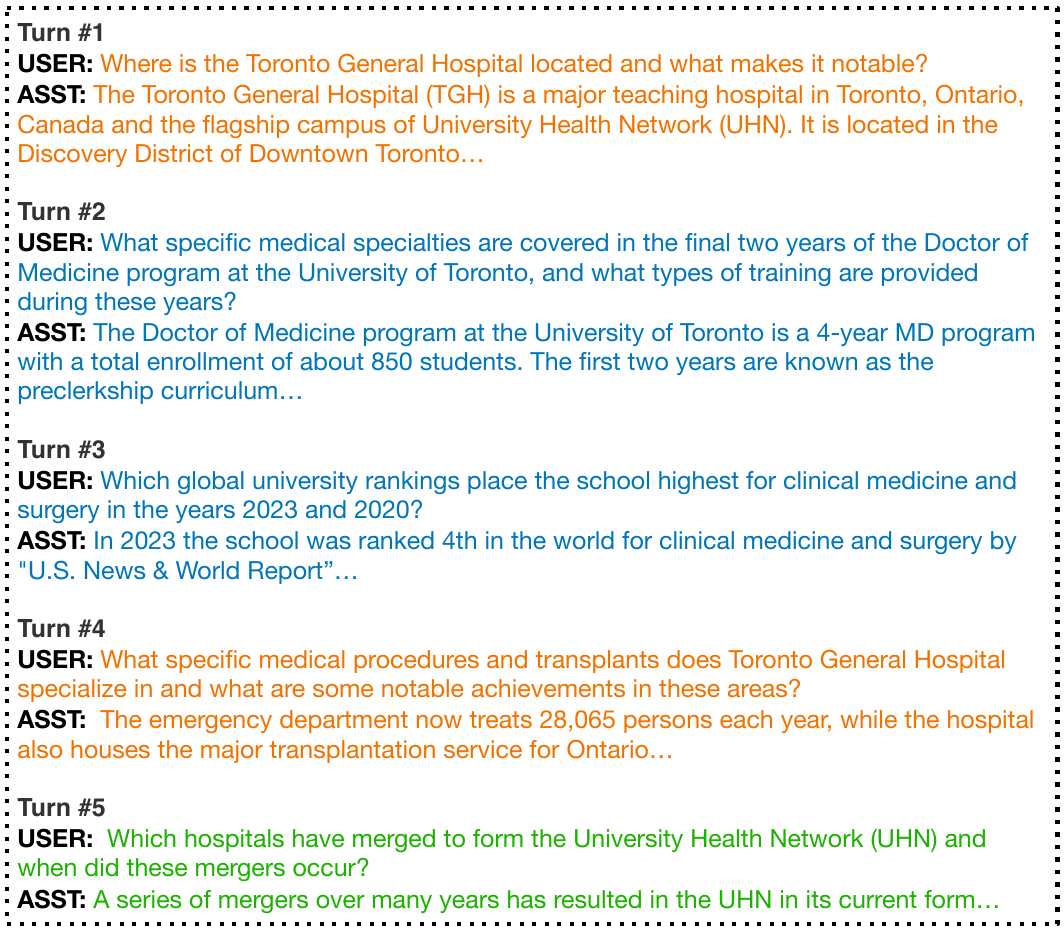}}
    \caption{First 5 turns of a \textit{DocTalk} conversation based on 3 Wikipedia articles: ``Toronto General Hospital'' (in \textcolor{orange}{orange}), ``University of Toronto Faculty of Medicine'' (in \textcolor{bluegray}{blue}), and ``United Health Network'' (in \textcolor{green}{green}).}
    \label{fig:doctalk}
\end{figure}

Our pipeline is applied to English Wikipedia documents, resulting in a conversational dataset named \textit{DocTalk}. In \textit{Stage 1}, we utilize the WIKIR toolkit \cite{frej-etal-2020-wikir}, providing query documents (i.e. anchor documents) and their relevance labels (qrels) indicating URL references to construct GDoc. We filter out documents with fewer than $10$ qrels to exclude shorter or obscure articles, resulting in a final set of 731,511 anchor documents. We sample up to $n=3$ documents per anchor document, including itself, and limit qrels to $20$ per document to form GDoc edges. In \textit{Stage 2}, documents are segmented by paragraph to reduce LLM-generated user utterances and hallucinations while utilizing all text segments. During \textit{Stage 3}, every vertex is visited once during traversal, resulting in a conversation with $m$ turns, where $m$ represents the total number of text segments. Corpus statistics are provided in Table \ref{tbl:corpus_stats}. Figure \ref{fig:doctalk} shows the first 5 turns of a \textit{DocTalk} conversation. The first 10 turns and an additional excerpt are in Appendix \ref{sec:doctalk-sample}.

\begin{table}
\centering
\caption{DocTalk statistics summary}
\label{tbl:corpus_stats}
\scalebox{0.8}{
\begin{tblr}{
  column{even} = {c},
  column{3} = {c},
  cell{1}{2} = {c=3}{},
  hline{1-3,7} = {-}{},
  stretch=0, colsep  = 3.0pt, rowsep=2pt
}
Total \# of conversations              & 730,707        &              &                 \\
\textbf{Metric}                        & \textbf{Mean} & \textbf{Std} & \textbf{Median} \\
\# turns per conversation              & 82.2          & 53.9         & 70              \\
Asst Utterance Length (\# words)       & 87.3          & 63.6         & 73              \\
User Utterance Length (\# words)       & 26.5          & 11.3         & 24              \\
\# Doc Shifts per conversation         & 23.3          & 15.9         & 19              
\end{tblr}}
\end{table}

While \textit{DocTalk} conversations are more direct and focused--omitting pleasantries and filler words--they effectively capture topical shifts and coherence necessary for effective human-LLM interactions. The primary aim of our conversation synthesis pipeline is to generate a large-scale dataset that fundamentally improves context memory and understanding in multi-turn conversations during pre-training, emphasizing structural aspects such as topic transitions over stylistic naturalness. Although this approach produces conversations that are stylistically less human-like, these aspects can be refined during post-training phases or through fine-tuning. Future work could entail extending our approach to synthesize more natural multi-turn dialogues on a larger scale.


\noindent\textbf{Human Evaluation} To ensure the quality of \emph{DocTalk}, we conducted a human evaluation as a form of quality control. We engaged 5 annotators to review 50 randomly selected dialogues from \emph{DocTalk} and assess their quality through a questionnaire. The questionnaire assessed various dimensions of the dialogues, including the contextual relevance of responses (Q1), the specificity and accuracy of user utterances (Q2), the amount of linguistic errors or level of grammatical correctness (Q3), the naturalness and overall resemblance to human-LLM conversation (Q4), the thematic connection between topics (Q5), and the logical consistency of topic transitions (Q6). For further comparison, we also evaluated 50 dialogues from WikiDialog using the same questionnaire (excluding Q5 and Q6 which are specific to DocTalk). See Appendix \ref{sec:questionnaire} for further details.

Human evaluation results are presented in Table \ref{tab:dataset_evaluation}. These scores confirm that the assistant utterances in \emph{DocTalk} are highly contextually relevant to the corresponding user utterances, the user utterances exhibit strong precision and specificity, and the dialogues maintain grammatical correctness and naturalness. Additionally, the high scores for Q5 and Q6 indicate that the Wikipedia articles selected in Stage 1 are semantically related and thematically cohesive, and that the topic shifts introduced in Stage 2 support logical and coherent multi-topic conversations. WikiDialog, sharing a common basis in Wikipedia articles, showed comparable levels of naturalness. However, \emph{DocTalk} excelled in contextual relevance, specificity, and grammatical accuracy, highlighting the effectiveness of our pipeline.

\begin{table}
\centering
\caption{Human evaluation for DocTalk and WikiDialog. Inter-annotator agreements are via Krippendorff’s $\alpha$}
\label{tab:dataset_evaluation}
\scalebox{0.8}{
\begin{tblr}{
  column{even} = {c},
  column{3} = {c},
  hline{1-2,8} = {-}{},
  stretch=0, colsep  = 3.0pt, rowsep=2pt
}
\textbf{Metric}           & $\alpha$ & \textbf{DocTalk} & \textbf{WikiDialog} \\
Q1. Contextual Relevance  & 0.87                  & 0.94             & 0.87                \\
Q2. Specificity           & 0.95                  & 0.96             & 0.64                \\
Q3. Linguistic Errors     & 0.94                  & 0.98             & 0.78                \\
Q4. Naturalness           & 0.80                  & 0.79             & 0.79                \\
Q5. Topic Relatedness     & 0.91                  & 0.95             & N/A                 \\
Q6. Topic Shift Coherence & 0.88                  & 0.83             & N/A                 
\end{tblr}}
\end{table}

\section{Experiments}

We utilize the annealing data quality assessment approach widely used in other studies \cite{blakeney2406does,grattafiori2024llama}. We continue pre-train the Mistral-7B-v0.3\footnote{https://huggingface.co/mistralai/Mistral-7B-v0.3} model for 8,000 steps (30B Tokens), applying a learning rate schedule that annealed from 3e-5 to 3e-6. We keep some learning rate bandwidth for instruction fine-tuning using ShareGPT data to prepare the model for benchmarking. For all our experiments, we construct a dataset comprising 25\% conversational data, 5\% Project Gutenberg books\footnote{Books data in the RedPajama dataset was replaced with Project Gutenberg content due to copyright concerns.}, and 70\%  RedPajama v1\footnote{Content from sites listed in the U.S. Notorious Markets for Counterfeiting and Piracy report were excluded.} Web content. Following this data composition, we prepare four variations to study our proposed solution in practice. 



\noindent\textbf{DocTalk}: We use the \emph{DocTalk} as described in Section \ref{sec:doctalk}. Along our main treatment experiment, we also prepare two ablation studies to examine DocTalk's constituent components: 1) We only applying \textit{Stage 1} and \textit{3} of the pipeline (w/o \textit{Stage 2}) to investigate the impact of GDial. In other words, we continuing pre-training on a conversational dataset curated solely with GDoc, preserving the document graph's original traversal sequence without segment reordering. 2) We only apply \textit{Stage 1} of the pipeline (w/o \textit{Stage 2} \& \textit{3}). Likewise, this baseline uses Wikipedia articles ordered solely by GDoc, excluding all user utterances.

\noindent\textbf{DocTalk*}: We continue pre-training on a \textit{DocTalk} variant restricted to the first 30 turns (roughly one-third of each conversation) to investigate if initial conversational turns are of higher quality than later ones. To match token counts across dataset configurations, we triplicate the resulting dataset.

\noindent\textbf{Plain Wiki}: Continue pre-training on Wikipedia articles without any additional treatment. Neither GDoc nor GDial are used.

\noindent\textbf{WikiDialog}: Continue pre-training on WikiDialog (triplicated to balance token count).

\subsection{Evaluation}

\subsubsection{Context Memory \& Understanding}

Existing popular multi-turn benchmarks, such as MTBench \cite{zheng2023judgingllmasajudgemtbenchchatbot} and WildBench \cite{lin2024wildbench}, evaluate high-level conversational reasoning. While these benchmarks overlap with our evaluation objectives, they do not explicitly assess context memory and understanding--MTBench is restricted to two turns, whereas WildBench features test samples of up to four turns, with only 20\% of dialogues exceeding two. We conduct a targeted evaluation by leveraging CoQA and devising a novel LLM-as-a-judge metric.


\noindent\textbf{CoQA} Similar to \citet{huber2024codiconversationaldistillationgrounded}, we investigate the turn-level performance of the baselines by leveraging the test set provided in the CoQA corpus \cite{adlakha2022topiocqaopendomainconversationalquestion}. Each sample consists of a text passage and a series of questions and answers that form a dialogue. At each turn, the LLM generates a response based on the passage and dialogue history. From the second turn onward, queries depend on both the passage and preceding dialogue, requiring the LLM to track context and resolve pronoun references (e.g., "Who is she?" or "How many were there?"). We report turn-level and overall F1, precision, and recall, computed via word overlap with reference responses.

\noindent\textbf{LLM-as-a-Judge} For our LLM-as-a-judge metric, we curtate a 73-sample multi-turn conversational dataset centered on shopping interactions. Each sample includes a conversation, a shopping-specific system prompt, and 1–3 manually labeled user intents that the response must address. Generating a response that addresses these intents would demonstrate an LLM’s ability to resolve anaphora and understand the relationship between instructions and inputs. Hence, improvements on this benchmark would reflect gains in context memory and understanding. 

Following the system prompt's instruction,  each response contains an introduction paragraph and a bullet-point paragraph. We measure how effectively the response addresses the user's intents via 4 metrics derived via an LLM-as-judge approach: $Intro$ for coverage in the introductory paragraph, $Bullet Point$ for coverage in bullet points, $Loose$ for coverage by either the introduction or bullet points, and $Strict$ for coverage by both. See Appendix \ref{sec:llm-as-a-judge} for a detailed explanation. 



\subsubsection{Guardrail Metrics}
To ensure that pre-training on \emph{DocTalk} does not degrade general LLM capabilities, standard benchmarks were used to evaluate knowledge, reasoning, and long-context capabilities of the base model (i.e., before IFT). For general knowledge, 5-shot MMLU\cite{hendrycks2021measuringmassivemultitasklanguage} is used. Commonsense reasoning is assessed using 25-shot ARC \cite{clark2018thinksolvedquestionanswering}, 5-shot WinoGrande(Wino.G)\cite{sakaguchi2019winograndeadversarialwinogradschema}, 0-shot PIQA\cite{bisk2019piqareasoningphysicalcommonsense}, and 10-shot HellaSwag(H.Swag)\cite{zellers-etal-2019-hellaswag}; logic reasoning relies on 4-shot MATH\cite{hendrycks2021measuringmathematicalproblemsolving}, 3-shot BBH\cite{suzgun2022challengingbigbenchtaskschainofthought}, and 5-shot GSM8k\cite{cobbe2021trainingverifierssolvemath}; fluency is evaluated with WikiText2\cite{merity2016pointersentinelmixturemodels}; and long-context performance with MuSiQue\cite{trivedi2022musiquemultihopquestionssinglehop} and 2WikiMultiHopQA(2WikiQA)\cite{ho-etal-2020-constructing}.

\section{Results \& Discussion}

\begin{table}
\centering
\caption{CoQA evaluation results}
\label{tab:coqa_metrics}
\scalebox{0.8}{
\begin{tblr}{
  column{even} = {c},
  column{3} = {c},
  column{5} = {c},
  hline{1-2,6,8} = {-}{},
  stretch=0, colsep  = 3.0pt, rowsep=2pt
}
\textbf{Model}                   & \textbf{F1}   & \textbf{Precision} & \textbf{Recall} & \textbf{Ave \# words} \\
\textbf{DocTalk}                 & 0.38 & 0.36      & 0.6             & 9.52                  \\
- w/o Stage 2 \& 3 & 0.29          & 0.27               & \textbf{0.61}   & 13.97                 \\
- w/o Stage 2                    & 0.27          & 0.25               & 0.52            & 12.06                 \\
\textbf{\textbf{DocTalk*}}       & \textbf{0.40}           & \textbf{0.37}               & \textbf{0.61}   & 9.07                  \\
\textbf{Plain Wiki}              & 0.34          & 0.31               & 0.6             & 12.58                 \\
\textbf{WikiDialog}              & 0.36          & 0.32               & 0.59            & 11.02                 
\end{tblr}}
\end{table}

\begin{table}
\centering
\caption{LLM-as-a-judge evaluation results}
\label{tbl:cc_metrics}
\scalebox{0.8}{
\begin{tblr}{
  column{even} = {c},
  column{3} = {c},
  column{5} = {c},
  hline{1-2,6,8} = {-}{},
    stretch=0, colsep  = 3.0pt, rowsep=2pt
}
\textbf{}                        & \textbf{Intro} & \textbf{Bullet Point} & \textbf{Loose} & \textbf{Strict} \\
\textbf{DocTalk}                 & \textbf{0.424} & 0.507                 & \textbf{0.542} & \textbf{0.331}  \\
- w/o Stage 2 \& 3 & 0.288          & 0.452                 & 0.381          & 0.271           \\
- w/o Stage 2                    & 0.263          & 0.522                 & 0.441          & 0.254           \\
\textbf{DocTalk*}                & 0.195          & \textbf{0.523}        & 0.525          & 0.186           \\
\textbf{Plain Wiki}              & 0.305          & 0.518                 & 0.424          & 0.254           \\
\textbf{WikiDialog}              & 0.271          & 0.482                 & 0.441          & 0.263           
\end{tblr}}
\end{table}

\noindent\textbf{Context Memory \& Understanding} CoQA turn-level and overall F1, precision, and recall scores are shown in Figure \ref{fig:coqa_graph} and Table \ref{tab:coqa_metrics}, respectively. Samples are provided in Appendix \ref{fig:coqa_sample}. Overall, the baseline pre-trained on \textbf{DocTalk*} and \textbf{DocTalk} outperformed all other baselines in terms of F1 and precision. Qualitatively, we observe that LLMs pre-trained on \textbf{Plain Wiki} or \textbf{WikiDialog} are more likely to generate erroneous responses, including incorrect or illogical answers, particularly in conversations with a larger number of turns. In contrast, LLMs pre-trained on \textbf{DocTalk} or \textbf{DocTalk*} demonstrate a significantly higher rate of accurate answers. It should also be highlighted that the responses generated by baselines pre-trained on \textbf{DocTalk*} and \textbf{DocTalk} demonstrated gains on F1 and precision despite being shorter in length. 

Furthermore, the baseline pre-trained on \textbf{DocTalk*} consistently outperformed all other baselines in precision and F1 score after the 8th turn, even surpassing the baseline pre-trained on \textbf{DocTalk}. We hypothesize that \textbf{DocTalk} dialogues become more abrupt and erratic in later stages as they exhaust all vertices of the dialogue graph and every segment of the Wikipedia document, leading to increasingly dissonant text in subsequent turns. By extracting the first 30 turns, we capture the dialogue's more coherent early sections with logical topic transitions. Future work could explore optimizing performance by adjusting the number of turns extracted. Regarding recall, most baselines showed similar performance, likely because LLMs tend to generate longer responses with greater overlap, boosting recall but reducing precision. The relatively poor performance of baselines pre-trained on \textbf{DocTalk} w/o Stage 2 and \textbf{DocTalk} w/o Stage 2 \& 3 underscores the importance of the dialogue graph and CR model for context memory and understanding.

Additionally, according to the LLM-as-a-judge metrics in Table \ref{tbl:cc_metrics}, pre-training on \textbf{DocTalk} and \textbf{DocTalk*} generally yield the best performance. The performance boost from incorporating the dialogue graph highlights its significance. We notice that the baseline pretrained on \textbf{DocTalk} achieves far greater Intro Coverage relative to all other baselines, including \textbf{DocTalk*}, which also results in a higher Overall strict coverage score. We hypothesize that longer pre-training dialogues, with more context understanding demonstrations across multiple turns, contribute to this improvement.

\begin{figure}[]
    \centering
    \scalebox{0.4}{
    \includegraphics[]{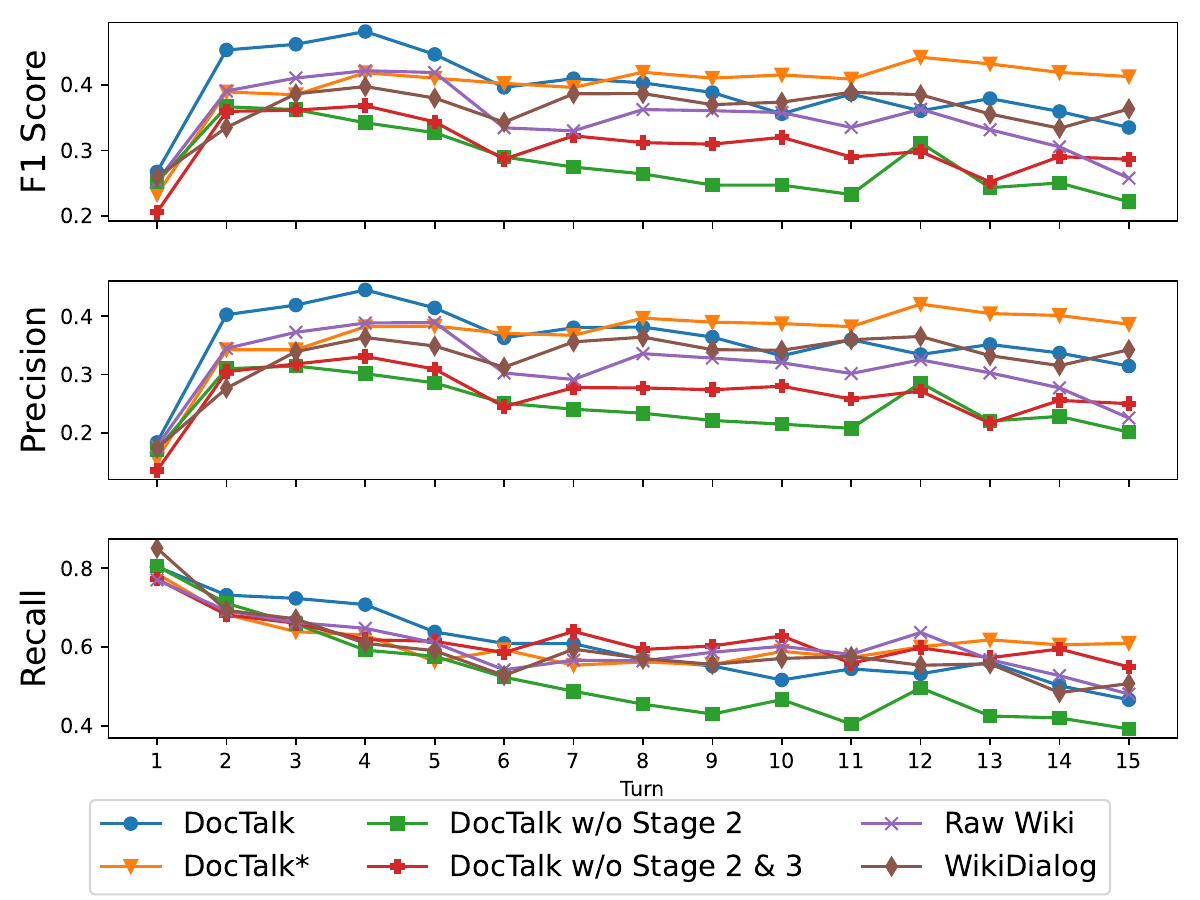}}
    \caption{Turn-level F1, precision, and recall plots.}
    \label{fig:coqa_graph}
\end{figure}

\begin{table*}
\centering
\caption{Guardrail benchmarking results for the latest checkpoint.}
\label{tbl:llm_eval}
\scalebox{0.8}{
\begin{tblr}{
  column{even} = {c},
  column{3} = {c},
  column{5} = {c},
  column{7} = {c},
  column{9} = {c},
  column{11} = {c},
  cell{1}{3} = {c=4}{},
  cell{1}{7} = {c=3}{},
  cell{1}{11} = {c=2}{},
  vline{2-4,7,8,10,11} = {1}{},
  vline{2-3,7,10-11} = {2-8}{},
  hline{1,3,7,9} = {-}{},
  stretch=0, colsep  = 3.0pt, rowsep=2pt
}
                    & \textbf{Knowledge} & \textbf{Commonsense Reasoning} &                 &                 &               & \textbf{Logical Reasoning} &               &                & \textbf{Fluency}  & \textbf{Long Context} &                  \\
Model               & \textbf{MMLU}      & \textbf{ARC}                   & \textbf{Wino.G} & \textbf{H.Swag} & \textbf{PIQA} & \textbf{MATH}              & \textbf{BBH}  & \textbf{GSM8k} & \textbf{WikiText} & \textbf{MuSiQue}      & \textbf{2WikiQA} \\
\textbf{DocTalk}    & 0.53               & 0.45                           & \textbf{0.71}   & 0.77            & \textbf{0.80} & \textbf{0.07}              & 0.49          & 0.20           & 6.04              & \textbf{0.12}         & \textbf{0.28}    \\
- w/o Stage 2  3    & \textbf{0.55}      & 0.44                           & \textbf{0.71}   & 0.77            & 0.78          & 0.06                       & \textbf{0.50} & 0.20           & 4.98              & 0.07                  & 0.18             \\
- w/o Stage 2       & \textbf{0.55}      & 0.46                           & 0.68            & 0.77            & \textbf{0.80} & 0.05                       & 0.46          & 0.19           & 5.74              & 0.05                  & 0.20             \\
\textbf{DocTalk*}   & 0.52               & 0.46                           & \textbf{0.71}   & 0.78            & 0.79          & 0.06                       & 0.46          & 0.12           & 6.45              & 0.06                  & 0.19             \\
\textbf{Plain Wiki} & \textbf{0.55}      & 0.45                           & 0.70            & \textbf{0.80}   & 0.79          & 0.05                       & 0.47          & 0.21           & \textbf{4.96}     & 0.08                  & 0.25             \\
\textbf{WikiDialog} & 0.51               & \textbf{0.48}                  & \textbf{0.71}   & 0.78            & \textbf{0.80} & \textbf{0.07}              & 0.49          & \textbf{0.22}  & 7.36              & 0.11                  & 0.21             
\end{tblr}}
\end{table*}

\begin{figure*}[]
    \centering
    \scalebox{0.43}{
    \includegraphics[]{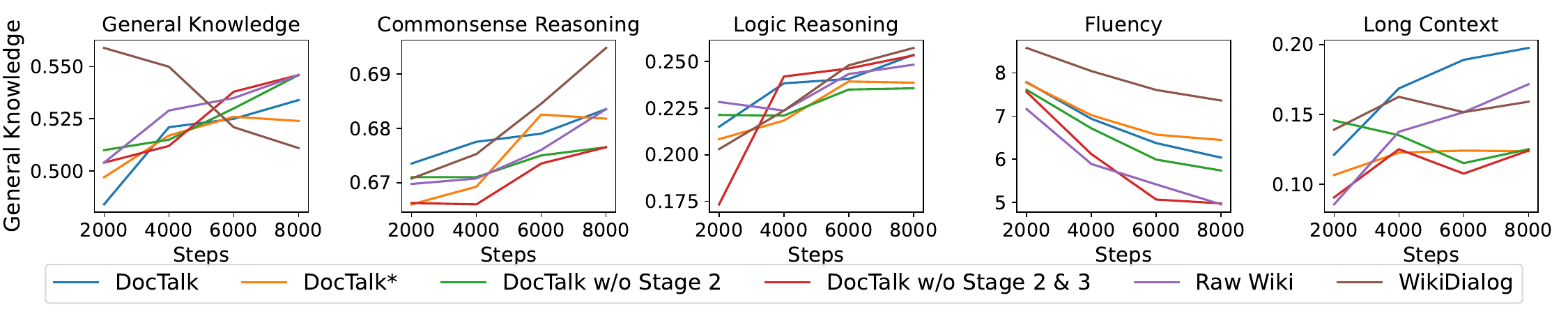}}
    \caption{Average general knowledge, reasoning, fluency, and long context scores evaluated every 2000 steps.}
    \label{fig:perf_plots}
\end{figure*}

\noindent\textbf{Guardrail Evaluation} Table \ref{tbl:llm_eval} presents the guardrail benchmarking scores, and Figure \ref{fig:perf_plots} depicts the scores during continued pre-training (before IFT). Overall, except for WikiText2 and MMLU, we observed no significant decline in base-model performance following continued pre-training on \textit{DocTalk}. The implemented baselines demonstrate comparable scores across general knowledge and reasoning tasks, showing steady improvement as the number of pre-training steps increases. For WikiText2, the perplexity reflects how much the curated dataset differs from the original Wikipedia. The baseline trained on \textbf{Raw Wiki} achieved the lowest perplexity, while \textbf{WikiDialog} attained the highest. This is expected since WikiText2 is a language modeling benchmark based on original Wikipedia articles; thus, the more the pre-training dataset deviates from Wikipedia, the poorer the performance.

We note that WikiDialog demonstrates  superiority in commonsense reasoning benchmarks, notably in ARC scores, compared to \textit{DocTalk}. This improvement can be attributed to WikiDialog's sentence-level granularity in question-answer pairs, which better align with the structure of reasoning benchmarks. However, we observe a degradation in MMLU performance for \textbf{WikiDialog} as training steps increase. This decline may be attributed to the large number of user utterances generated via Dialogue Inpainting, showing adverse impact.

The baseline pre-trained on \textbf{DocTalk} showed significant performance gains on MuSiQue and 2WikiHotPotQA. We hypothesize that the long multi-turn conversations in \emph{DocTalk} serves to enhance the LLM's long-context capabilities. In contrast, baselines pre-trained on \textbf{DocTalk} w/o Stage 2 or w/o Stage 2 \& 3 applied performed worse than the baselines pre-trained on \textbf{Raw Wiki}, emphasizing the dialogue graph's effectiveness in enhancing long-context performance.

\section{Related Work}
\noindent\textbf{LLM Pre-training} Pre-training is crucial stage where LLMs acquire foundational knowledge, typically through unsupervised learning with language modeling or masked language modeling objectives on large-scale web-crawled text \cite{Radford2019LanguageMA}. Typically, data selection--encompassing filtering, deduplication, and mixing--is performed to curate a high-quality pre-training dataset \cite{albalak2024surveydataselectionlanguage}. Sophisticated curation methods have been proposed to align pre-training data with downstream tasks. Previous work has explored incorporating task-specific instructions and responses \cite{cheng2024instructionpretraininglanguagemodels}, merging related documents to facilitate cross-document reasoning \cite{shi2024incontextpretraininglanguagemodeling}, introducing soft prompts \cite{gu-etal-2022-ppt}, and using regex patterns to format reading comprehension data \cite{jiang2024improvingdomainadaptationextendedtext}.

\noindent\textbf{Information Seeking Dialogue Synthesis} There has been significant research devoted to the synthesis of information seeking dialogue. A common approach involves iteratively generating document-grounded conversations by extracting relevant information and crafting coherent responses to questions posed by human annotators \cite{kim2022generatinginformationseekingconversationsunlabeled, hwang-etal-2023-dialogizer, wu-etal-2023-inscit}. This approach, while effective, relies heavily on resource-intensive human annotation. To mitigate this, alternatives like AutoConv \cite{li-etal-2023-autoconv}, Dialogue Inpainting \cite{dai2022dialoginpaintingturningdocuments} and SOLID\cite{askari-etal-2025-solid} leverage LLMs instead. AutoConv fine-tunes LLMs on human dialogues using iterative prompting to generate user and system responses, Dialogue Inpainting inserts conversational utterances between document sentences, and SOLID uses a self-seeding, multi-intent self-instructing approach. In addition to the broad information seeking paradigm, prompt-based LLM approaches have also been used to synthesize dialogues for specific use-cases such as education \cite{wang-etal-2024-book2dial} and emotional support \cite{seo-lee-2024-diagesc}.

\section{Conclusion}
We present a conversational data synthesis pipeline that transforms web-crawled text into multi-turn, multi-topic dialogues, enhancing LLMs' context memory and understanding during pre-training. Using this pipeline, we curate \textit{DocTalk}, the largest conversational dataset to date. Future enhancements could refine the CR model by scoring transitions with full dialogue context and extend the pipeline to general texts via semantic matching for document relatedness. Future investigations could explore using these dialogues for grounded synthetic conversation generation, potentially enhancing conversational naturalness in data subsets for mid-training or post-training. Additionally, the \emph{DocTalk} corpus mainly consists of straightforward question-and-answer exchanges. Future work could explore incorporating more natural conversational dynamics, such as clarifications, misconceptions, requests for further details, or supporting evidence, to enhance the naturalness of the dialogues. It would be interesting to examine how these enhancements affect multi-turn conversational abilities. 


\section{Limitations}
Similar to other projects that involve the use of synthetic data, there is an inherent risk of hallucination. This risk can have adverse effects on the performance and reliability of the LLM. For our pipeline, we have taken steps to mitigate this risk by only synthesizing the user utterances, not the assistant utterances. By doing so, we aim to reduce the likelihood of hallucinations negatively influencing the LLM's behavior. However, despite these precautions, it is important to recognize that the possibility of hallucination cannot be entirely eliminated. We plan to release the data only to be used for research purpose, and derivatives of data should not be used outside of research contexts.


\bibliography{custom}

\clearpage
\appendix

\onecolumn

\section{Appendix}

\subsection{Cost Advantages}
\label{sec:cost}

\emph{DocTalk} consists of approximately 8 billion tokens. If every token in the corpus were synthetically generated using GPT-4’s pricing (2.50 USD per 1 million input tokens and 10.00 USD per 1 million output tokens), the cost of generating \emph{DocTalk} would amount to 80,000 USD. This estimate excludes the cost of input tokens, which would vary depending on the complexity of the prompts used. By leveraging our pipeline, only the user queries are generated synthetically, while the assistant’s responses are used to prompt the LLM to generate the user queries. This approach significantly reduces costs. Specifically, generating 500 million tokens for user queries would cost approximately 5,000 USD for output tokens and 18,750 USD for input tokens, totaling 23,750 USD. This represents nearly a 70\% reduction in cost compared to the fully synthetic generation approach.

\subsection{LLM-as-a-Judge Evaluation}
\label{sec:llm-as-a-judge}
In order to rigorously evaluate how well our model retains contextual information, understands user instructions, and addresses specific intents within a multi-turn conversation, we define a set of metrics designed to assess the model’s ability to follow the system prompt, resolve anaphora, and cover the user’s requested topics. In particular, we focus on the extent to which both the introduction paragraph and the bullet-point paragraph of the model’s response address the core user intents identified in our curated shopping-related dataset. By quantifying the proportion of these required intents satisfied in either or both of these components, we aim to capture the model’s degree of context memory and understanding in a manner that is both comprehensive and interpretable. Overall, evaluation results show a 90\% consistency rate with human annotations, demonstrating the reliability of our metrics.

\subsubsection{Response Structure}

Following the system prompt's instruction, each response (denoted as $resp$) produced by the LLM consists of two components:
\begin{itemize}
    \item An \textbf{introduction paragraph} (denoted as $resp\_intro$).
    \item A \textbf{bullet-point paragraph} (denoted as $resp\_bulletpoint$).
\end{itemize}

\subsubsection{Metrics for Measuring Context Memory and Understanding}

Let $intent$ denote the set of user intents that must be addressed, and let $|intent|$ represent the total number of these required user intents. We define:
\begin{equation}
A_{intro} = \{\text{user intents addressed by the introduction}\}
\end{equation}
\begin{equation}
A_{bulletpoint} = \{\text{user intents addressed by the bullet points}\}
\end{equation}

The function:
\begin{equation}
intent\_coverage(A, intent) = \frac{|A \cap intent|}{|intent|}
\end{equation}

measures the proportion of required user intents successfully covered by a given set $A$.

We define four evaluation metrics as follows:

\begin{itemize}
    \item \textbf{Intro}: The $intro$ metric assesses if the response introduction addresses each of the user intents.
    \begin{equation}
        intro(resp\_intro, requirement) = intent\_coverage(A_{intro}, intent)
    \end{equation}

    \item \textbf{Bullet Point}: The $bulletpoint$ metric assess if the bullet points in the response addresses the user intents.
    \begin{equation}
    bulletpoint(resp\_bulletpoint, requirement) = intent\_coverage(A_{bulletpoint}, intent).
    \end{equation}

    \item \textbf{Loose}:
    The $loose$ metric checks if \emph{either} the introduction \emph{or} the bullet points address the user intents. In set-theoretic terms, this corresponds to the union of addressed intents:
    \begin{equation}
    loose(resp, intent) = intent\_coverage(A_{intro} \cup A_{bulletpoint}, intent).
    \end{equation}

    \item \textbf{Strict}:
    The $strict$ metric requires that \emph{both} the introduction \emph{and} the bullet points address the user intents. This corresponds to the intersection of addressed intents:
    \begin{equation}
    strict(resp, intent) = intent\_coverage(A_{intro} \cap A_{bulletpoint}, intent).
    \end{equation}
\end{itemize}

\subsubsection{LLM-as-a-Judge Approach}

To compute these metrics, we adopt an LLM-as-a-judge approach:
\begin{itemize}
    \item We select 10 samples from the curated 73-sample dataset to elicit responses from a LLM(Claude 3).
    \item These responses are then manually labeled with binary indicators to specify whether each user intent was addressed or not, forming the sets $A_{intro}$ and $A_{bulletpoint}$ for each response.
    \item The manually labeled responses serve as a ground truth for fine-tuning the LLM-as-a-judge prompt.
\end{itemize}

\begin{figure}[]
    \centering
    \scalebox{0.4}{
    \includegraphics[]{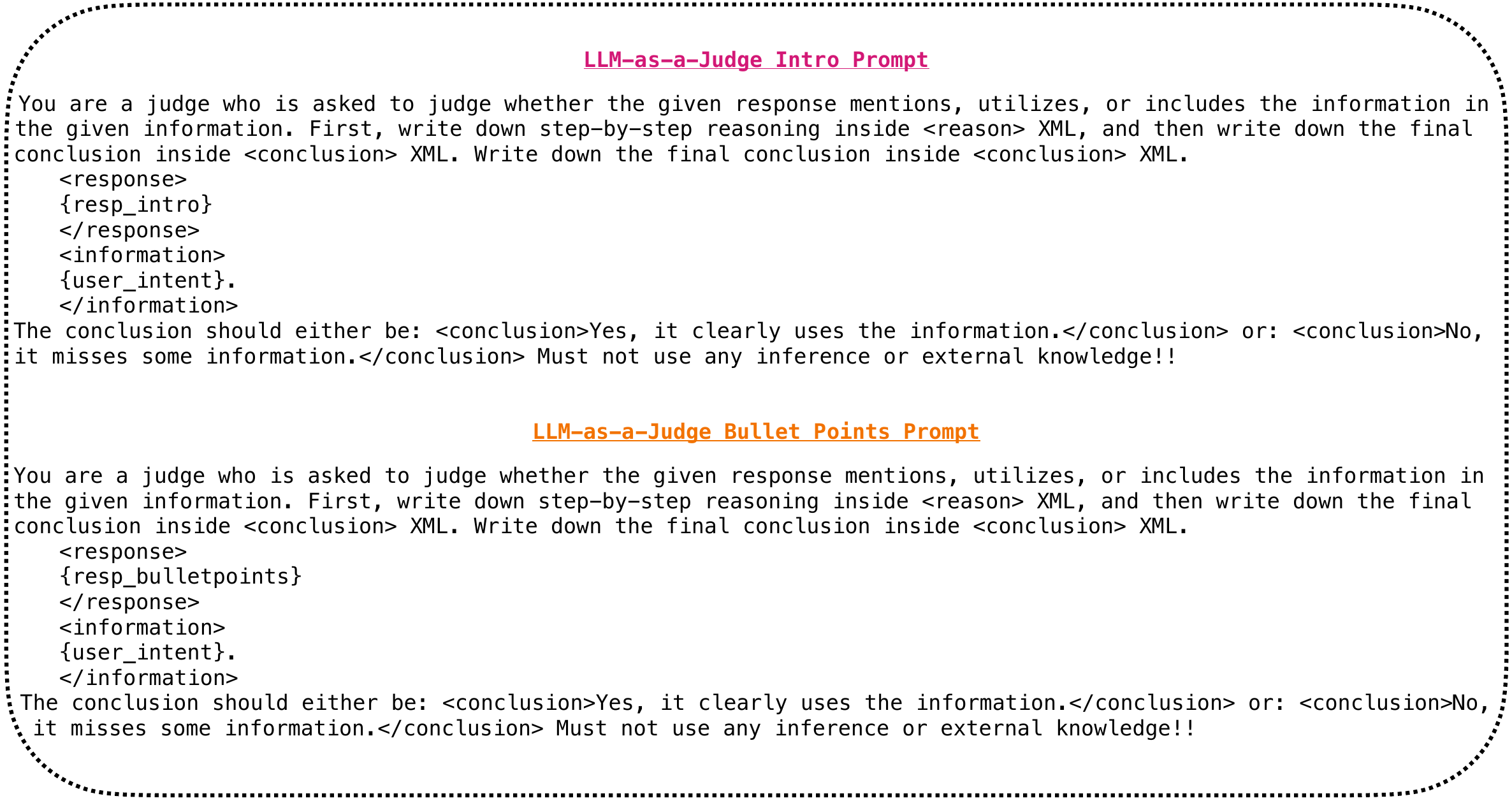}}
    \caption{Sample from the LLM-as-a-judge evaluation dataset.}
    \label{fig:judge_prompt}
\end{figure}

By refining the judge prompt with these 10 labeled samples, we achieve at least 80\% recall and precision in evaluating whether the LLM's responses demonstrate context memory and understanding. Thus, the LLM-as-a-judge can reliably estimate $A_{intro}$, $A_{bulletpoint}$, and consequently $intent\_coverage$. The prompt templates are provided in Fig \ref{fig:judge_prompt}.

\begin{figure}[]
    \centering
    \scalebox{0.4}{
    \includegraphics[]{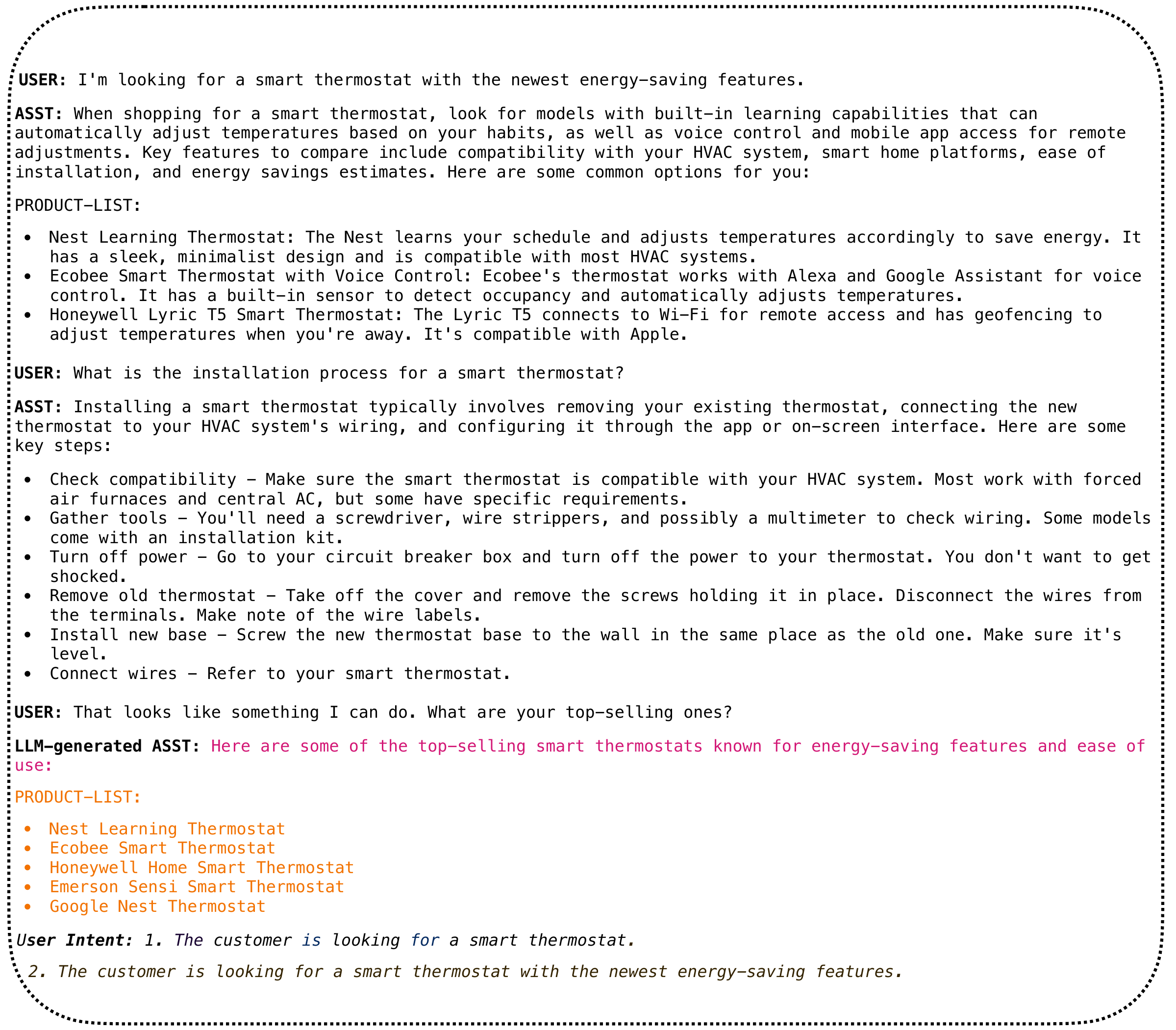}}
    \caption{Sample from the LLM-as-a-judge evaluation dataset.}
    \label{fig:judge_sample}
\end{figure}

The final scores shown in Table \ref{tbl:cc_metrics} are obtained by averaging the scores across all samples. Figure \ref{fig:judge_sample} provides a representative sample from the evaluation dataset.

\subsection{User Utterance Generation Prompt}
\label{sec:user-utt-prompt}
\begin{figure}[]
    \centering
    \scalebox{0.65}{
    \includegraphics[]{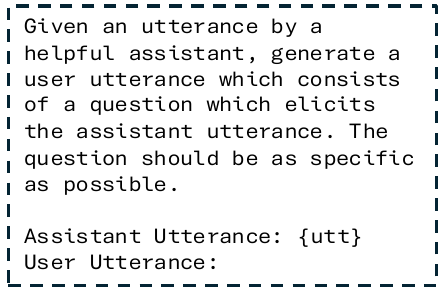}}
    \caption{Prompt template for user utterance generation in Stage 3.}
    \label{fig:prompt_temp}
\end{figure}

The prompt used for user utterance in Stage 3 is provided Figure \ref{fig:prompt_temp}.

\subsection{Document \& Dialogue Graph Traversal Algorithms}
\label{sec:algos}

The algorithms for sampling the document graph in Stage 1 and the dialogue graph in Stage 2 is provided in Algorithm \ref{alg:doc_sampling} and \ref{alg:dial_sampling}, respectively.

\begin{algorithm*}
\caption{Document Graph traversal in Stage 1}
\label{alg:doc_sampling}
\begin{algorithmic}
\REQUIRE Directed, weighted document graph $G_{doc} = (V_{doc}, E_{doc})$, number of documents $n$, start vertex $v_{\text{start}}$, out-degree centrality $deg^{\text{-}}(v)$ for all $v \in V_{doc}$
\ENSURE A set of $n$ related documents $D$

\STATE $D \gets \{v_{\text{start}}\}$
\STATE $v_{\text{current}} \gets v_{\text{start}}$

\FOR{$k \gets 2$ to $n$}
    \STATE Let $O_{v_{\text{current}}} = \{ v_j \mid (v_{\text{current}} \rightarrow v_j) \in E_{doc} \}$
    \IF{$O_{v_{\text{current}}} = \emptyset$}
        \STATE \textbf{break} \COMMENT{No outgoing edges; cannot sample further}
    \ENDIF

    \FORALL{$v_j \in O_{v_{\text{current}}}$}
        \STATE $p_j \gets \frac{deg^{\text{-}}(v_j)}{\sum_{v_{u} \in O_{v_{\text{current}}}} deg^{\text{-}}(v_{u})}$
    \ENDFOR

    \STATE \text{Sample $v_{\text{next}}$ from $O_{v_{\text{current}}}$ according to probabilities $\{p_j\}$}
    \STATE $D \gets D \cup \{v_{\text{next}}\}$
    \STATE $v_{\text{current}} \gets v_{\text{next}}$
\ENDFOR

\RETURN $D$
\end{algorithmic}
\end{algorithm*}

\begin{algorithm*}
\caption{Dialogue Graph traversal in Stage 2}
\label{alg:dial_sampling}
\begin{algorithmic}
\REQUIRE Directed dialogue graph $G_{dial} = (V_{dial}, E_{dial})$, number of utterances $m$, start vertex $v_{\text{start}}$, CR scores $r_{(u \rightarrow v)}$ for all edges $(u \rightarrow v) \in E_{dial}$
\ENSURE A sequence of $m$ Assistant utterances $U$

\STATE $U \gets \{v_{\text{start}}\}$
\STATE $v_{\text{current}} \gets v_{\text{start}}$
\STATE $V_{\text{visited}} \gets \{v_{\text{start}}\}$

\FOR{$t \gets 2$ to $m$}
    \STATE Let $O_{v_{\text{current}}} = \{ v_j \mid (v_{\text{current}} \rightarrow v_j) \in E_{dial} \}$
    \STATE $O_{v_{\text{current}}} \gets O_{v_{\text{current}}} \setminus V_{\text{visited}}$ \COMMENT{Prune previously visited vertices}

    \IF{$O_{v_{\text{current}}} = \emptyset$}
        \STATE \textbf{break} \COMMENT{No more valid next utterances}
    \ENDIF

    \FORALL{$v_j \in O_{v_{\text{current}}}$}
        \STATE $r_j \gets r_{(v_{\text{current}} \rightarrow v_j)}$
    \ENDFOR

    \STATE Let $Z = \sum_{v_j \in O_{v_{\text{current}}}} r_j$
    \FORALL{$v_j \in O_{v_{\text{current}}}$}
        \STATE $p_j \gets \frac{r_j}{Z}$
    \ENDFOR

    \STATE \text{Sample $v_{\text{next}}$ from $O_{v_{\text{current}}}$ with probability distribution $\{p_j\}$}

    \STATE $U \gets U \cup \{v_{\text{next}}\}$
    \STATE $V_{\text{visited}} \gets V_{\text{visited}} \cup \{v_{\text{next}}\}$
    \STATE $v_{\text{current}} \gets v_{\text{next}}$
\ENDFOR

\RETURN $U$

\end{algorithmic}
\end{algorithm*}

\subsection{DocTalk Sample}
\label{sec:doctalk-sample}
\begin{figure}[]
    \centering
    \scalebox{0.5}{
    \includegraphics[]{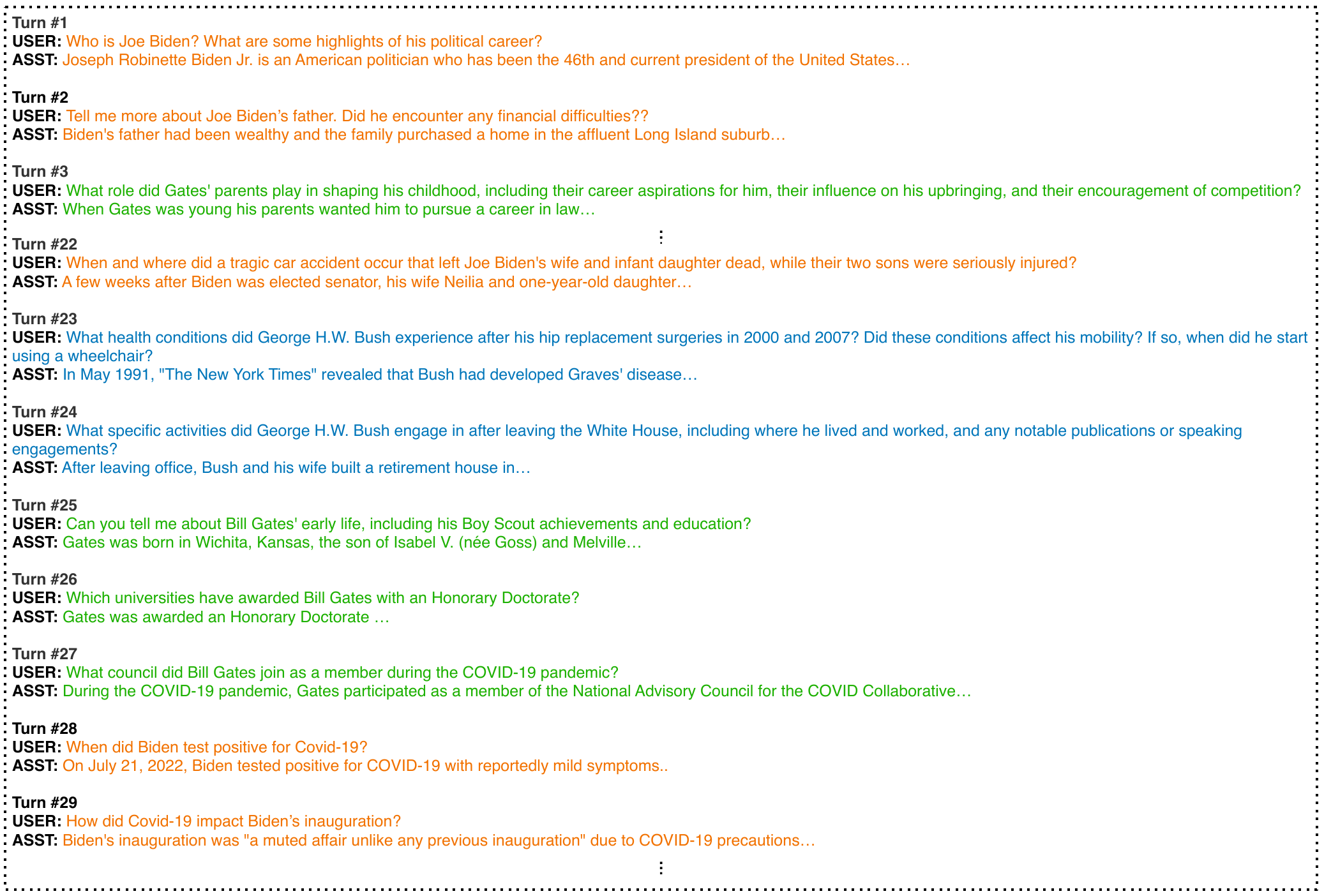}}
    \caption{Excerpt of a \textit{DocTalk} conversation based on 3 Wikipedia articles: `Joe Biden'' (in \textcolor{orange}{orange}), ``George H.W. Bush'' (in \textcolor{bluegray}{blue}), and ``Bill Gates'' (in \textcolor{green}{green}). }
    \label{fig:doctalk-sample-1}
\end{figure}

\begin{figure}[]
    \centering
    \scalebox{0.5}{
    \includegraphics[]{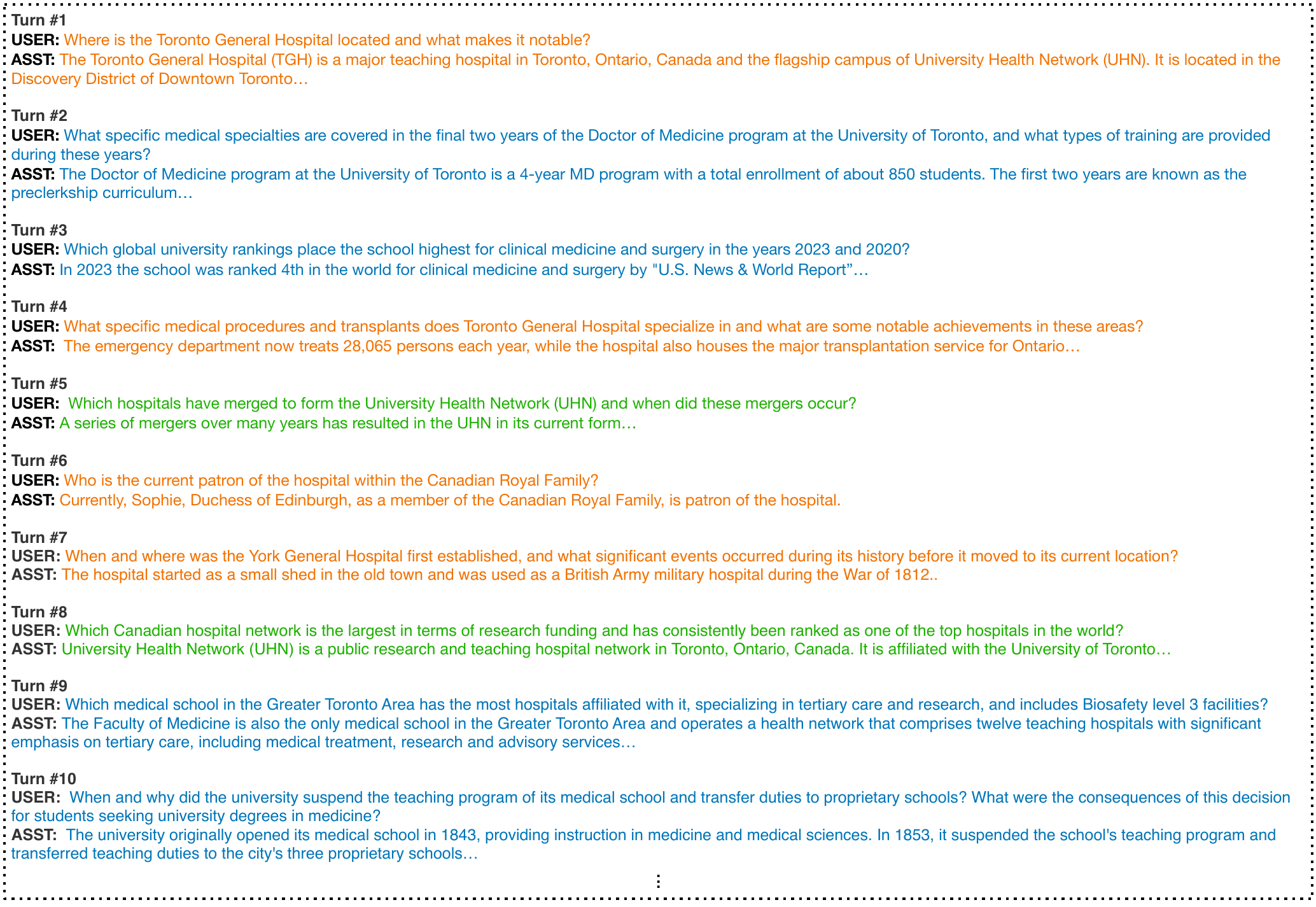}}
    \caption{First 10 turns of a \textit{DocTalk} conversation based on 3 Wikipedia articles: ``Toronto General Hospital'' (in \textcolor{orange}{orange}), ``University of Toronto Faculty of Medicine'' (in \textcolor{bluegray}{blue}), and ``United Health Network'' (in \textcolor{green}{green}).}
    \label{fig:doctalk-sample-2}
\end{figure}

Samples from \emph{DocTalk} are provided in Figure \ref{fig:doctalk-sample-1} and Figure \ref{fig:doctalk-sample-2}, where different colors represent utterances sourced from distinct documents.

\subsection{CoQA Generation Sample}
\label{sec:coqa-sample}
\begin{figure}[]
    \centering
    \scalebox{0.4}{
    \includegraphics[]{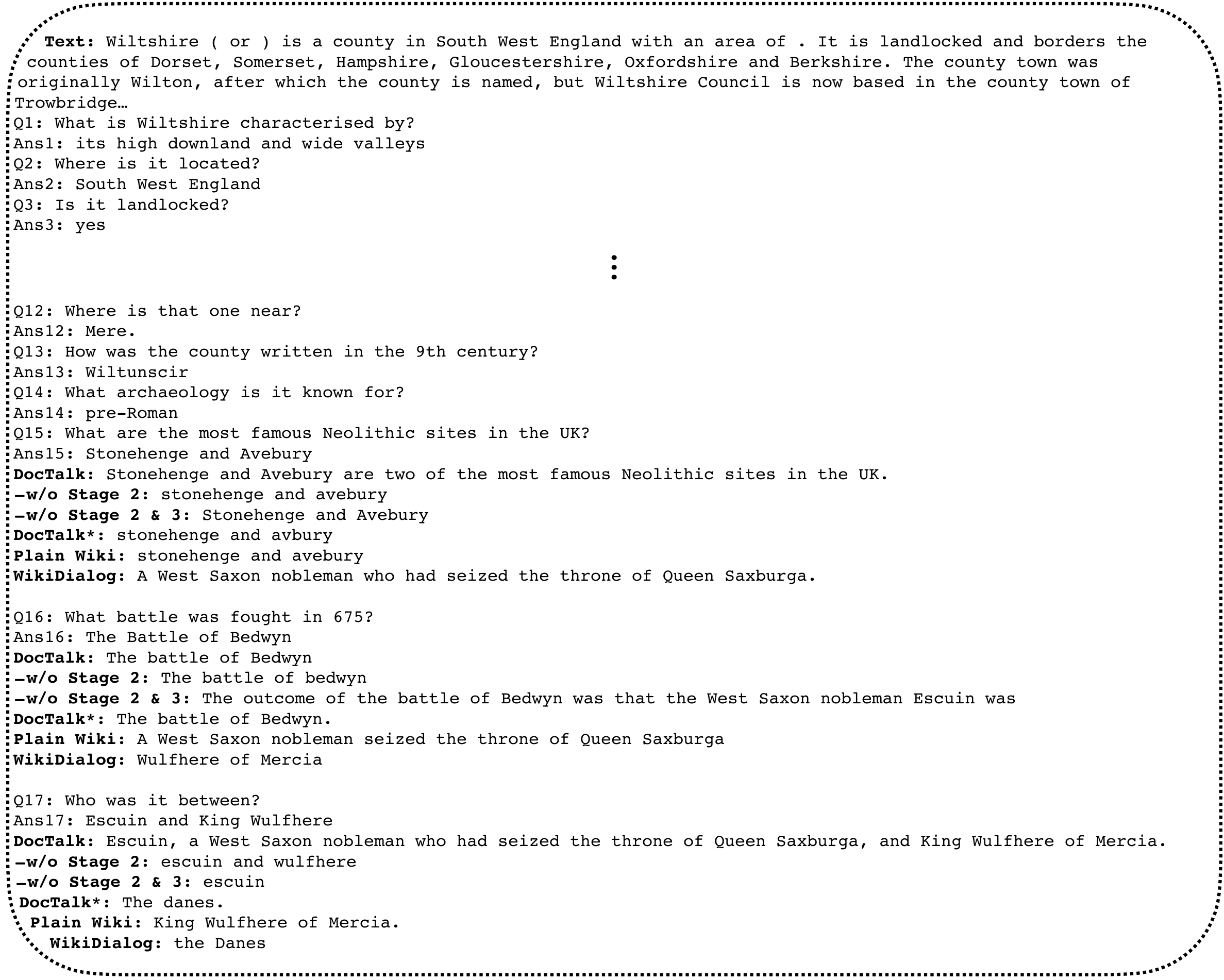}}
    \caption{Samples of responses generated during the CoQA evaluation. Each example displays the outputs from each baseline model, starting from the 15th turn.}
    \label{fig:coqa_sample}
\end{figure}

A sample response from CoQA evaluation is provided in Figure \ref{fig:coqa_sample}. Starting from the 15th turn, we provide the responses generated by each of the implemented baselines. 

\subsection{Conversational Reward Model}
\label{sec:cr_model}
To finetune the CR model,  we construct the training dataset using open-source multi-turn dialogue datasets, prioritizing information-seeking corpora like HybriDialogue \cite{nakamura-etal-2022-hybridialogue}, InScitic \cite{wu-etal-2023-inscit}, and TopiOCQA \cite{adlakha2022topiocqaopendomainconversationalquestion}. However, while semantically relevant, assistant utterances from these datasets are much shorter than the utterances in our dataset, typically only containing the direct answer to the user’s question. Additionally, only a small portion of conversations in these datasets contain a sufficient number of dialogue turns. Hence, we supplement the information-seeking datasets with more general dialogue datasets such as ShareGPT\footnote{The ShareGPT dataset: https://sharegpt.com/} and UltraChat \cite{ding2023enhancingchatlanguagemodels}. Additionally, we also include the Wizard of Wikipedia (WoW) corpus \cite{dinan2019wizardwikipediaknowledgepoweredconversational}, a knowledge-grounded open-domain dialogue dataset, which emphasizes conversational naturalness. The final training set includes 2000 conversations each from ShareGPT and UltraChat, and all suitable conversations from HybriDialogue, InScitic, WoW, and TopiOCQA (2240 in total). For evaluation, we extract 100 conversations from each dataset’s test set. We use Mean Reciprocal Rank (MRR) to evaluate our CR model performance. MRR measures the effectiveness of ranking by calculating the average reciprocal rank of the first relevant result across all queries.

\subsection{Human Evaluation}
\label{sec:questionnaire}
\begin{figure}[]
    \centering
    \scalebox{0.4}{
    \includegraphics[]{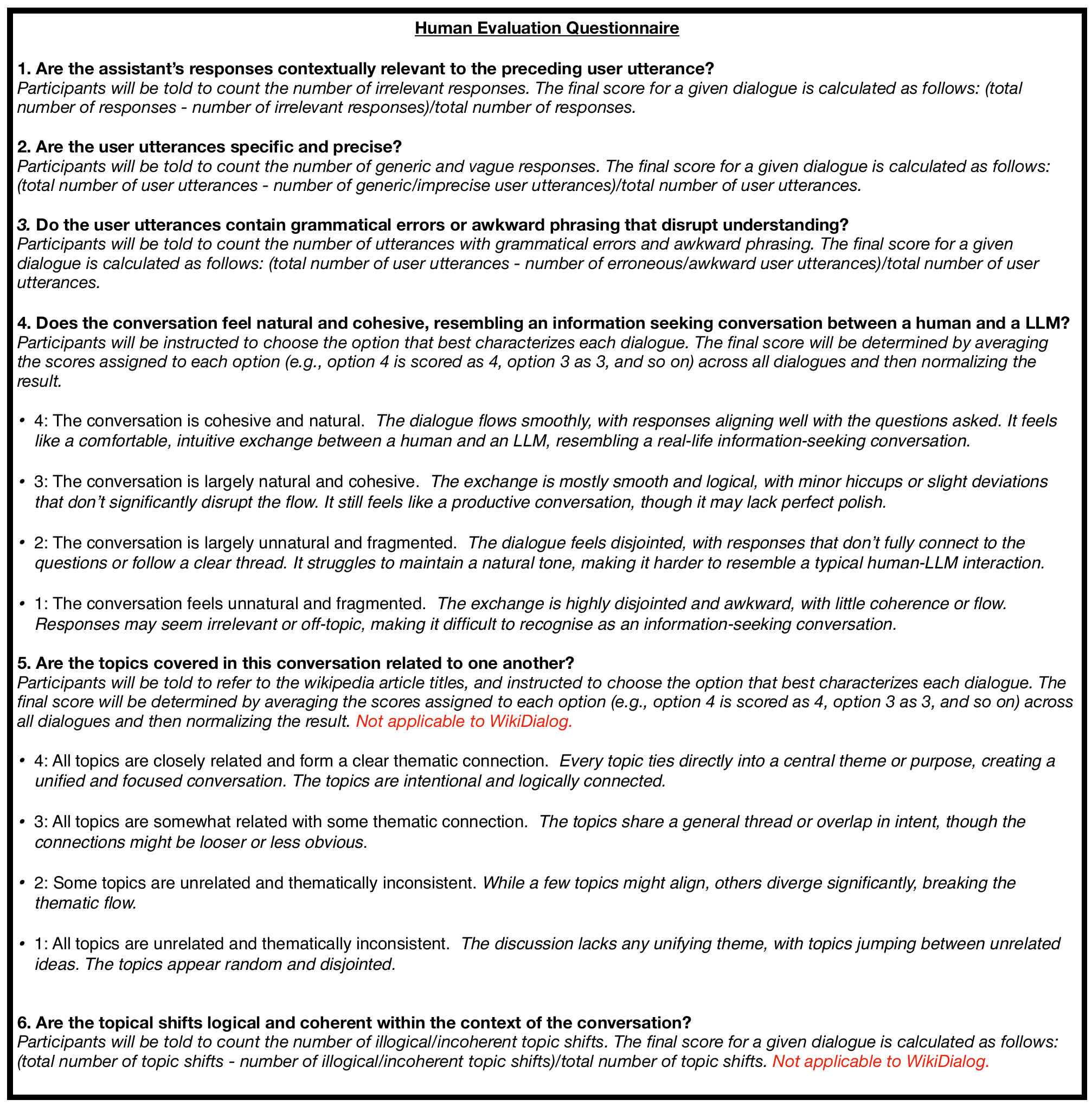}}
    \caption{Questionnaire used for human evaluation. The questionnaire assesses dialogue quality through six criteria: Q1. contextual relevance of the assistant's responses, scored by counting irrelevant responses; Q2. specificity and precision of user utterances, measured by tallying generic/vague instances; Q3. grammatical correctness and phrasing clarity, evaluated by counting errors or awkwardness; Q4. naturalness and overall resemblance to human-LLM conversation, rated on a 4-point scale; Q5. thematic consistency of topics, scored on a 4-point scale, with reference to Wikipedia articles (not applicable to WikiDialog); and Q6. logical coherence of topic shifts, calculated by counting illogical transitions.}
    \label{fig:human_eval_questions}
\end{figure}

For the human evaluation, we engaged 5 native English-speaking annotators. Each annotator was presented with 50 conversations from \emph{DocTalk} and 50 conversations from WikiDialog. The evaluation was conducted in person in two 1.5 hour sessions. The annotators evaluated the dialogues using a detailed questionnaire, provided in Fig. \ref{fig:human_eval_questions}, designed to measure various aspects of conversational quality. First, they assessed the contextual relevance of the assistant’s responses by identifying irrelevant replies and calculating a relevance score based on the proportion of relevant responses to the total number of responses. Next, they evaluated the specificity and precision of user utterances by determining the proportion of specific responses relative to the total. Grammatical accuracy and phrasing were also examined, with annotators counting user utterances containing errors or awkward phrasing and computing the proportion of grammatically accurate responses relative to the total. In addition, the naturalness and cohesiveness of each conversation were rated on a scale from 1 to 4. For \emph{DocTalk} conversations, annotators examined topical relatedness by referencing Wikipedia article titles, categorizing the strength of thematic connections (on a scale from 1 to 4), and assessed the logical coherence of topic shifts by calculating the proportion of logical shifts out of all transitions. This multifaceted approach provided a comprehensive evaluation of conversational dynamics across both datasets.

\end{document}